\documentclass[11pt,a4paper]{article}
\usepackage[utf8]{inputenc}
\usepackage[T1]{fontenc}
\usepackage{lmodern}
\usepackage[margin=24mm]{geometry}
\usepackage{amsmath,amssymb}
\usepackage{graphicx,booktabs,array,microtype}
\usepackage{xurl}
\usepackage[hidelinks]{hyperref}
\title{Intervention anchors and scientific verification in synthetic vascular predictive representations}
\author{}
\date{}
\begin{document}
\maketitle
\vspace{-2.5em}
\begin{center}
Lingsen You\textsuperscript{1,2}, Yujun Guo\textsuperscript{3}, Xinyu Zhong\textsuperscript{4}, Zisu Peng\textsuperscript{1,2}, Wentong Wang\textsuperscript{1,2}, Li Shen\textsuperscript{1,2*}, Junbo Ge\textsuperscript{1,2*}
\end{center}
{\footnotesize
\textsuperscript{1}Department of Cardiology, Zhongshan Hospital, Fudan University, Shanghai Institute of Cardiovascular Diseases, State Key Laboratory of Cardiovascular Diseases, NHC Key Laboratory of Ischemic Heart Diseases, National Clinical Research Center for Interventional Medicine, Key Laboratory of Viral Heart Diseases, Chinese Academy of Medical Sciences, Shanghai 200032, China.\par
\textsuperscript{2}Institutes of Biomedical Sciences, Fudan University, Shanghai 200030, China.\par
\textsuperscript{3}College of Biomedical Engineering, Fudan University, Shanghai 200438, China.\par
\textsuperscript{4}Department of Cardiology, Shidong Hospital, Yangpu District, Shanghai 200438, China.\par
* Correspondence: Li Shen (shen\_li@fudan.edu.cn); Junbo Ge (jbge@zs-hospital.sh.cn).\par
}
\begin{center}Technical Note -- 6 October 2026\end{center}
\begin{abstract}
Complete orthogonal predictive coordinates do not by themselves bind a latent direction to a named intervention. We present a mathematical and synthetic audit motivated by vascular device--vessel suitcordance. Capacity-matched least-squares predictors were exactly equivalent under complete fixed output transforms, whereas an anchor-only observer recovered interpretations only within the span of known perturbation signatures. Six three-dimensional configurations across 64 seeds gave a maximum paired prediction discrepancy of 6.7 \ensuremath{\times} 10\ensuremath{^{-15}} but a median untransported edit error of 1.513. Coordinate transport removed that error. Noisy and weak anchors constrained calibration stability, and changing the representation basis required recalibration or verified transport. Across 256 additional fits in dimensions 3--24, prediction equivalence persisted within 4.0 \ensuremath{\times} 10\ensuremath{^{-15}}. We then evaluated nine deliberate runnable fault classes across 64 seeds. All 576 faulty executions completed, but each violated at least one reconstruction, prediction, delivered-edit or scope contract; all 320 valid control records passed. Repeating a faulty implementation gave exact self-agreement despite error against the separately computed simulator expectation. For one omitted-direction defect, probe coverage followed its analytic law, and rank-aware abstention protected unsupported interpretations. Scalar-noise experiments exposed both missed weak faults and excessive rejection under narrow relative tolerances. These controls provide an executable separation of prediction, semantic support and scientific acceptance. They are synthetic numerical audits, not clinical validation, neural JEPA-Anything replication, agent learning or patient treatment-effect estimation.
\end{abstract}
\section{Introduction}
Vascular digital twins must connect observed vessel states to interpretable responses if they are to support device selection or follow-up. Device--vessel suitcordance provides one motivating vocabulary: mechanical, cellular, and physicochemical--immune ecology describe complementary aspects of a device interacting with a vessel \cite{ref1}. Imaging-based vascular digital twins offer another motivation for connecting measurements and longitudinal predictions \cite{ref2}. These frameworks suggest useful scientific questions, but their labels cannot by themselves validate coordinates learned by a predictive model.

JEPA-Anything introduces orthogonal predictive factorization with learned projectors, factor-specific prediction, and state synthesis across multiple domains \cite{ref3}. Its predictive and geometric objectives motivate examining the interpretation of a complete set of coordinates. The authors explicitly distinguish factor geometry from causal meaning. Here, we study the exact orthogonal linear boundary, where prediction-preserving changes of coordinates can be analyzed without neural optimization. Our study is an independent control audit rather than a reproduction of the reported neural results.

Related work makes verification central to scientific AI. ScienceIDE packages executable scientific acceptance rules \cite{ref10}; ScienceBuddy connects procedural revision and model updates \cite{ref11}; Discovery Foundation Models distinguish validated knowledge progress from future discovery capability \cite{ref12}. We use these ideas to audit the acceptance rules themselves in a small mathematical system, without implementing their agents or training procedures.

The clinical distinction between prediction and intervention is consequential. In the nationwide Biomarker Research cohort of patients with acute myocardial infarction and Killip class III or IV, intra-aortic balloon pump (IABP) use was associated with higher in-hospital mortality; the investigators emphasized residual confounding and an observational interpretation \cite{ref4}. This association cannot be assigned as a causal action label for a digital twin. We cite the cohort as a clinical motivation for preserving evidential boundaries and do not access its patient data or calibrate a simulator to its estimates.

Representation identifiability depends on observations and assumptions \cite{ref5}. Temporal structure or known intervention targets can supply additional information under specified conditions \cite{ref6,ref7}. We restrict an observer to paired perturbation signatures in named and latent spaces. The audit asks what such anchors identify, how noise and weak signatures affect stability, and which executable checks distinguish a valid prediction from an unsupported named edit. We combine algebraic controls, ambiguity witnesses, dimensional stress tests and an explicitly constructed fault bank (Figure 1).

\section{Methods}
\subsection{Study scope and information boundary}
This is an original mathematical and synthetic computational study. The three state coordinates are dimensionless conceptual analogues of mechanical (M), cellular (C), and physicochemical--immune (I) ecology. They are neither biomarkers nor experimentally measured vascular mechanisms. Six arbitrary linear configurations, labeled A--F, vary the persistence of these coordinates; they are not six validated vascular beds. No patients, images, tissue assays, clinical outcomes or device measurements were analyzed.

The simulator evaluator knows the synthesis matrix Q and named state x to score errors. The anchor-only observer receives A and B, defined below, but cannot use Q, additional named state--latent pairs, transition eigenstructure, or external biological supervision. This restriction is essential: if Q or sufficiently informative paired states are available, coordinate transport is already identified. We do not claim that factors are unidentifiable from every possible longitudinal dataset.

\subsection{Complete orthogonal coordinates and matched prediction}
Column-vector notation is used throughout. Q is a 3 \ensuremath{\times} 3 orthogonal synthesis matrix, allowed to have determinant +1 or \ensuremath{-}1. Analysis and synthesis are

\begin{equation}
u=Q^{\mathsf T}x,\qquad x=Qu,\qquad Q^{\mathsf T}Q=QQ^{\mathsf T}=I.
\tag{1}\label{eq:1}
\end{equation}
For a row-wise input design D and output matrix Y, monolithic ordinary least squares gives W = D\ensuremath{^{\dagger}}Y. Three independent output heads use the same seven predictors, consisting of three state coordinates, three action coordinates and an intercept. They regress the columns of YQ, giving V = D\ensuremath{^{\dagger}}(YQ). Thus, for any test design Z,

\begin{equation}
V=WQ,\qquad (ZV)Q^{\mathsf T}=ZW.
\tag{2}\label{eq:2}
\end{equation}
The parameter count is 21 for each predictor. This equality checks a fixed complete output transformation, not learned nonlinear projectors, optimization, activity regularization, or independent nonlinear branches.

Orthogonality also gives \ensuremath{\lVert Q^{\mathsf T}e\rVert_2^2=\lVert e\rVert_2^2} for any prediction residual e. A latent coordinate edit \ensuremath{\delta u=e_j} produces \ensuremath{\delta x=Qe_j}; a specified named-state edit \ensuremath{\delta x=e_j} requires \ensuremath{\delta u=Q^{\mathsf T}e_j}. We report Euclidean error for a single edit and RMS error across three edits, \ensuremath{\lVert Q-I\rVert_F/\sqrt{3}}. Coordinate labels and state edits are distinct from treatment effects.

\subsection{Synthetic prediction configurations}
Each configuration follows \ensuremath{y=F_rx+a+\varepsilon}, with x drawn independently from N(0,I), action a, and observation noise \ensuremath{\varepsilon} drawn from N(0,\ensuremath{0.05^2I}). For r = 0,\ldots{},5, the diagonal of Fr is (0.62 + 0.035r, 0.77 + 0.02r, 0.90 + 0.01r). The only nonzero off-diagonal entries are \ensuremath{F_{12}=0.012}, \ensuremath{F_{21}=0.008}, \ensuremath{F_{23}=0.008} and \ensuremath{F_{32}=0.010}. These stable matrices have arbitrary time units and no physiological calibration.

For each of 64 seeds and each configuration, 1,024 training observations contain one uniformly selected coordinate action of magnitude +1 or \ensuremath{-}1. The 512 independent test observations contain two distinct simultaneous coordinate actions with independent signs. Both models use exactly the same observations and least-squares convention. Mean squared error is averaged over observations and output coordinates. We record the largest paired prediction difference across all 384 configuration--seed fits. Q is drawn by QR decomposition of a Gaussian matrix with diagonal-sign correction and is shared by the six configurations within a seed.

An exact rotation sweep uses angles 0--180 degrees in the M--C plane and a fixed residual (0.2, \ensuremath{-}0.1, 0.3). A separate six-row isotropic tight frame, \ensuremath{\pm\sqrt{3}e_j}, has population covariance I; its factor standard deviations remain one under orthogonal transforms. The activity threshold 0.8 is illustrative. We also inspect an anisotropic tight frame with covariance diag(4,1,0.25). We do not treat coordinate activity as generally rotation-invariant.

\subsection{Anchors and supported interpretation}
Let A contain named-state perturbation signatures as columns, and B = Q\ensuremath{^{\mathsf T}}A their noiseless latent signatures. The anchor-only linear map from latent signatures to named states is T = AB\ensuremath{^{\dagger}}. Define P = AA\ensuremath{^{\dagger}}, the orthogonal projection onto span(A). For any state change d,

\begin{equation}
TQ^{\mathsf T}d=AA^{\dagger}d=Pd.
\tag{3}\label{eq:3}
\end{equation}
The complementary named-state residual \ensuremath{\lVert(I-P)d\rVert_2} identifies the component unsupported by these anchors. Conversely, B A\ensuremath{^{\dagger}}d transports a supported named edit into latent coordinates. A zero residual permits interpretation within the specified linear model; it does not establish a biological mechanism. Nonzero residuals trigger abstention from the unsupported component rather than a zero-effect claim.

For anchor rank k = 1,2,3 we use \ensuremath{A=[e_1,\ldots,e_k]}. With H = diag(1,\ldots{},1,\ensuremath{-}1,\ldots{},\ensuremath{-}1), where the first k entries are one, \ensuremath{Q'=HQ} has exactly the same anchors: \ensuremath{(Q')^{\mathsf T}A=Q^{\mathsf T}A}. Yet unanchored directions change sign for k < 3. H reverses two directions for rank one and reflects one direction for rank two, giving explicit ambiguity witnesses in \ensuremath{O(3)}. A known orientation constraint restricting the transformation to \ensuremath{SO(3)} can remove the remaining rank-two sign ambiguity; that prior is deliberately absent here.

\subsection{Noisy anchors and coordinate transfer}
For full-rank A = I, observed signatures are \ensuremath{\widetilde B=Q^{\mathsf T}A+\overline E}, with independent Gaussian noise standard deviations 0.05, 0.2 or 0.5. The noise is averaged over n = 1,4,16 or 64 measurements; increasing-n conditions are nested within each seed. We compare the unrestricted map A \ensuremath{\widetilde{B}}\ensuremath{^{\dagger}} with orthogonal Procrustes calibration \ensuremath{\widehat R=UV^{\mathsf T}}, where \ensuremath{U\Sigma V^{\mathsf T}} is the SVD of A \ensuremath{\widetilde{B}}\ensuremath{^{\mathsf T}} \cite{ref8}. Reflection is allowed. This second method uses the known orthogonal-map prior and is not expected to be suitable for arbitrary nonlinear encoders.

Calibration is evaluated on the three unit state edits, using \ensuremath{\lVert TQ^{\mathsf T}-I\rVert_F/\sqrt{3}} or \ensuremath{\lVert \widehat RQ^{\mathsf T}-I\rVert_F/\sqrt{3}}. For noise-free identity anchors these targets overlap the calibration directions, so the result is an exact reconstruction check rather than independent biological validation. Noisy calibration uses independent true, noiseless signatures for evaluation. Conditioning is tested with A = diag(1,s,s), s = 1,0.3,0.1,0.03,0.01, measurement-noise standard deviation 0.02 and 16 repeats. Both rank and singular-value scale are reported.

For coordinate transfer, independent source and target bases Qs and Qt encode the same named space. Applying a source map directly gives \ensuremath{Q_sQ_t^{\mathsf T}}; target calibration gives \ensuremath{Q_tQ_t^{\mathsf T}=I}. A known cross-basis transport also restores consistency. This is a test of representation-coordinate mismatch, not physiological domain shift or cross-hospital validation.

\subsection{Dimensional extension of prediction and anchor controls}
We extended the complete-coordinate prediction control to d = 3, 6, 12 and 24, with 64 simulation seeds per dimension. Each fit used 256 training observations and 128 independent test observations, with observation-noise standard deviation 0.05; sample counts were held fixed across dimensions. Within each replicate, the monolithic and transformed-output predictors used the same input design, observations and least-squares convention. The synthesis matrix Q was a complete orthogonal d \ensuremath{\times} d transform. We compared synthesized predictions directly, rather than treating agreement between loss summaries as sufficient. These dimensions are numerical control settings and do not represent additional measured vascular ecologies or validated physiological models.

The additional dimension generator used F = diag(linspace(0.65,0.93,d)), independent standard-normal states, one signed coordinate action at training and two distinct signed coordinate actions at testing. Both predictors used 2d + 1 input columns and d outputs, giving d(2d + 1) parameters. The expansion master seed was 2026100602, with separate SeedSequence streams for dimension, fault, probe and tolerance analyses.

For the coverage control, A consisted of \ensuremath{\lfloor}d/2\ensuremath{\rfloor} distinct unit-coordinate anchors and B = Q\ensuremath{^{\mathsf T}}A. We evaluated the recovered component of named perturbations and their complementary residual using P = AA\ensuremath{^{\dagger}}. The identity AB\ensuremath{^{\dagger}}Q\ensuremath{^{\mathsf T}} = P was checked in each dimension. Coverage refers to the declared linear anchor span; recovery within that span cannot validate an unanchored coordinate. Each dimension therefore supplies an algebraic extension of the original control, rather than a new biological dataset. Figure 5 reports replicate-level prediction discrepancies and supported-span checks.

\subsection{Executable scientific contracts and controlled faults}
Following the task-validity distinction in ScienceIDE \cite{ref10}, we constructed a finite, prespecified set of candidate pipelines with known mathematical behavior. Nine defective variants altered synthesis transposition, physical output scale, axis retention, transport of a named edit, reuse of a calibration from a different basis, the sign of a coordinate, action inclusion, output content, or the stated scope of anchor coverage. Five valid variants comprised the healthy reference pipeline, a consistently transported signed permutation, a consistently transported reflection, a consistent change of latent units, and an unchanged healthy pipeline. A separate variant raised an execution error. The design contains 15 variants across 64 seeds, giving 960 records: 896 runnable records, including 576 defective records and 320 valid controls, and 64 execution-error records. These are repeated evaluations of specified fault classes, not 960 independently discovered scientific tasks.

The fault-bank reference used F = diag(0.65,0.80,0.93) and 128 independent standard-normal states and actions per seed, with no observation noise or fitted-coefficient error. The reference prediction was computed directly as Fx + a. Reconstruction was tested on those sampled states. Reconstruction, prediction and actual delivered-edit contracts used maximum absolute discrepancy at most 10\ensuremath{^{-10}}; numerical anchor rank used the same absolute tolerance. Execution required finite outputs and the specified output shapes. Each faulty case was paired with a passing full-span reference replacement; that replacement is a validity witness, not a measured learner.

We applied five contracts: execution, coordinate reconstruction, noiseless state prediction, end-to-end recovery of a requested supported edit, and truthful reporting of supported scope. Execution established whether a result was produced under the declared interface; the other contracts tested its scientific meaning within the simulator. For an analysis map C and synthesis map D, the reconstruction check evaluated DC against the named-space identity. End-to-end edit errors were measured in named physical coordinates after synthesis, so a calibration residual computed in isolation could not establish that the requested edit was actually delivered. The scope check compared the declared dimension with the numerical anchor rank. For these coordinate-axis examples it rejects a full-span claim with missing anchors; it is not a complete verifier of every named direction in arbitrary anchor geometry. Applicable checks and numerical tolerances were fixed before this fault-bank execution and were not tuned to the observed pass rates.

The unit-conversion control used C = 2Q\ensuremath{^{\mathsf T}} and D = Q/2, so DC = I even though the latent coordinates were scaled. This valid control tests whether the contracts respect declared coordinate units; orthogonality of C alone was not imposed as a universal requirement. Correctly matched analysis and synthesis, named-space reconstruction, and the physical end-to-end edit metric were the relevant criteria. A faulty physical output gain was tested separately. Execution-error cases were recorded as such and excluded from the denominator of runnable semantic errors. Candidate repeatability was also separated from mathematical validity: calling the same faulty implementation twice gives regression agreement by construction while disagreeing with a separately computed named-space oracle. The oracle is independent of that injected fault, not an independent research group or independent source of biological evidence.

One fixed example records the candidate operation, observed contract residual, prescribed coordinate correction and repeated checks. The correction is a deterministic application of the known transport identity. The trace documents how the verified output changes after that correction; it does not measure an agent's ability to diagnose a defect or learn from feedback. Figure 6 presents contract outcomes and this correction trace; the shared-fault comparison is recorded in the seed-level result table.

\subsection{Probe coverage for an unanchored defect}
We examined a three-dimensional rank-two anchor set, A = [e1,e2], with a candidate that reversed only the third direction. The anchor observations are unchanged by this defect, so an anchor residual alone cannot distinguish the correct and defective extensions. We then sampled unit-axis probes uniformly with replacement from \{e1,e2,e3\}. For budgets n = 1, 2, 4, 8, 16 and 32, 4,096 Monte Carlo trials estimated the probability of observing at least one disagreement. The exact detection probability is 1 \ensuremath{-} (2/3)\ensuremath{^{n}} because each probe misses the affected axis with probability 2/3. This probability describes the specified probe mechanism, not sensitivity to arbitrary faults or clinically important changes.

A separate targeted control applies one probe in the unanchored complement. Such a probe supplies additional evaluation information; it is unavailable to an observer restricted to the original two anchors. We also distinguish fault detection from scope protection: the rank-two scope gate can withhold interpretation of the unsupported direction even when no probe has detected the sign defect. Figure 7 compares random-probe coverage with the analytic probability and targeted complement testing. The supported-span gate separately withholds the unsupported third direction.

\subsection{Tolerance behavior under controlled measurement noise}
To examine how a valid execution can receive an uninformative or misleading numerical verdict, we used a scalar reference response \ensuremath{\delta}, healthy measurements \ensuremath{\delta} + \ensuremath{\varepsilon}, and defective measurements \ensuremath{-}\ensuremath{\delta} + \ensuremath{\varepsilon}, with \ensuremath{\varepsilon} drawn from N(0,\ensuremath{\sigma}\ensuremath{^{2}}), \ensuremath{\sigma} = 10\ensuremath{^{-4}}, and \ensuremath{\delta} varied from 10\ensuremath{^{-6}} to 1. At each of 25 logarithmically spaced signal settings, 16,384 measurements per condition evaluated rejection when the absolute discrepancy from \ensuremath{\delta} exceeded a specified tolerance. We compared an absolute tolerance 5\ensuremath{\sigma}, a relative tolerance 0.05\ensuremath{\delta}, and a hybrid tolerance max(5\ensuremath{\sigma},0.05\ensuremath{\delta}). All three rules reused the same measurements, giving 819,200 scalar observations across conditions; their verdicts are paired policy evaluations. Healthy rejection is a false positive; defective rejection is a true positive for this particular sign-reversal fault.

The signal \ensuremath{\delta} is a known simulator response scale and is not a clinical effect size or threshold of medical importance. We reported both true-positive detection and healthy false-positive rates, rather than reporting detection alone. The analytic healthy false-positive probability is 2[1 \ensuremath{-} \ensuremath{\Phi}(\ensuremath{\tau}/\ensuremath{\sigma})], and the defective detection probability is \ensuremath{\Phi}((2\ensuremath{\delta} \ensuremath{-} \ensuremath{\tau})/\ensuremath{\sigma}) + 1 \ensuremath{-} \ensuremath{\Phi}((2\ensuremath{\delta} + \ensuremath{\tau})/\ensuremath{\sigma}), where \ensuremath{\tau} is the selected tolerance and \ensuremath{\Phi} is the standard-normal distribution function. These probabilities describe the declared Gaussian measurement model. Figure 8 displays the empirical rates alongside these analytic controls; finite trial variation is not uncertainty about a patient treatment effect.

\subsection{Expansion provenance and reporting}
The additional dimension, fault-bank, probe-budget and tolerance analyses were specified after inspection of the original study and its first computational results. They are recorded as an expansion addendum before their own execution, rather than presented as part of an external preregistration. Original results and their specifications are retained. Records identify the seed, configuration, candidate variant, contract, numerical residual, verdict and execution status where applicable. No new model weights, agent policy, user-feedback loop or patient-level source was evaluated. Comparisons therefore concern known coordinate transformations, selected defects and fixed verification rules.

\subsection{Reproducibility}
The local analysis specification was saved before the first execution, with master seed 20261006; it was not externally preregistered. An explicitly recorded post-execution addendum introduced identity and signed-permutation bases, coefficient transport, action-response transport, ten-step rollout controls and anisotropic activity sensitivity. The rollout control numerically composes the fitted maps for ten steps; it does not supply observed longitudinal trajectories. Original outcomes were retained, without selection or parameter retuning. Separate simulation streams cover prediction, span, noise, conditioning, transfer and the expansion experiments. Descriptive bands are 2.5th--97.5th percentiles across 64 seed replicates, not confidence intervals for patients or for an estimated biological effect. The 384 prediction fits share seeds within configurations, and are not 384 independent random-basis replicates. No significance testing was used.

Python 3.12.14, NumPy 2.3.5 and Matplotlib 3.11.2 were used. The arXiv ancillary archive supplies seeds, specifications, executable scripts, all seed-level result tables, execution traces and original figures. The scripts regenerate the 384 original observation datasets, 256 additional dimension datasets, 64 fault-bank input/output records, probe draws and scalar measurements.

\subsection{DECLARATION OF GENERATIVE AI AND AI-ASSISTED TECHNOLOGIES}
Codex assisted with LaTeX assembly, prose editing, and consistency checks. The authors reviewed and revised all AI-assisted text and take responsibility for the submitted manuscript.

\section{Results}
\subsection{Prediction was preserved while uncalibrated edits changed}
Across six configurations and 64 seeds, the median test MSE of the monolithic predictor was 0.002519. The maximum paired output discrepancy was 6.66 \ensuremath{\times} 10\ensuremath{^{-15}} and the maximum MSE discrepancy was 1.17 \ensuremath{\times} 10\ensuremath{^{-17}}. Additional identity, signed-permutation and random-basis controls yielded maximum coefficient-transport error 2.00 \ensuremath{\times} 10\ensuremath{^{-15}}, action-response discrepancy 2.22 \ensuremath{\times} 10\ensuremath{^{-15}}, and ten-step prediction discrepancy 4.80 \ensuremath{\times} 10\ensuremath{^{-14}}. These values reflect finite-precision implementation of Equation (2).

The angle sweep retained squared residual norm 0.14, while an uncalibrated unit M edit had error \ensuremath{2\sin(\theta/2)}, reaching two at 180 degrees (Figure 2). Exact transport restored the requested edit. Across 64 random orthogonal bases, the RMS error across three uncalibrated edits had median 1.513 and empirical 2.5th--97.5th percentiles 0.743--1.756; transported edit error was at most 6.91 \ensuremath{\times} 10\ensuremath{^{-16}}. The isotropic activity control remained fully active at every angle. In the anisotropic control, a 45-degree M--I rotation changed SDs from (2,1,0.5) to (1.458,1,1.458), reducing the illustrative activity penalty from 0.1 to zero. Activity therefore provided neither a universal invariance nor a semantic label.

\subsection{Anchor coverage and noise constrained interpretation}
Noiseless anchor maps recovered their supported columns within 6.66 \ensuremath{\times} 10\ensuremath{^{-16}}, and Equation (3) held within 7.83 \ensuremath{\times} 10\ensuremath{^{-16}}. Rank-one and rank-two coordinate-anchor sets supported one and two of the three unit edits, respectively; the remaining edits had unsupported residual one and were marked for abstention (Figure 3A). The alternative complement-sign bases produced exactly identical anchor signatures, demonstrating ambiguity in the omitted directions without relying on numerical completion.

At measurement-noise SD 0.2, median inverse-map error decreased from 0.324 to 0.039 with one versus 64 repeated measurements; corresponding Procrustes errors decreased from 0.172 to 0.019 (Table 1; Figure 3B--C). The comparison reflects a correctly specified orthogonal prior. At SD 0.5 with a single measurement, the inverse-map error had an upper empirical percentile of 9.824, showing a substantial noise tail that is retained in the reported results.

\subsection{Full rank and source calibration were insufficient controls}
Every conditioning configuration had true rank three. As its true condition number rose from one to 100, median inverse-map error increased from 0.0079 to 0.5860; Procrustes error increased from 0.0043 to 0.2033 (Figure 4A). Weak anchors therefore remained unstable despite algebraic coverage. Applying a source calibration to an independently rotated target basis gave median RMS error 1.487, with empirical percentiles 0.822--1.734. Target recalibration reduced the error below 9.74 \ensuremath{\times} 10\ensuremath{^{-16}}; known cross-basis transport also restored consistency below 1.04 \ensuremath{\times} 10\ensuremath{^{-15}} (Figure 4B).

\subsection{Complete-coordinate equivalence extended across dimensions}
Across d = 3, 6, 12 and 24 and 64 seeds per dimension, the largest discrepancy between monolithic and synthesized transformed-output predictions was 4.00 \ensuremath{\times} 10\ensuremath{^{-15}}. The largest supported-span identity residual was 8.88 \ensuremath{\times} 10\ensuremath{^{-16}}. The coordinate-anchor examples supported \ensuremath{\lfloor}d/2\ensuremath{\rfloor} of d orthogonal unit directions; the complementary directions remained outside the anchor span (Figure 5). The result extends the implementation check for a complete fixed output transform and does not establish that a learned nonlinear factorization has equivalent behavior.

\subsection{Execution and regression agreement did not certify the candidate pipelines}
All 896 runnable records produced outputs under the declared execution interface, while the 64 error-control records were recorded separately. Among the 576 specified defective records, 576 were rejected by at least one applicable scientific contract. The 320 valid records, including consistent unit conversion, were accepted by 320 all-contract evaluations (Figure 6A). Contract-level outcomes reveal which declared property each selected fault violates; these totals do not estimate the prevalence of errors in real research code.

Repeating the same faulty candidate produced exactly identical predictions, whereas the named-space oracle recorded maximum absolute prediction error 3.519 (seed-zero duplicate-gain control). The fixed correction trace reduced the maximum absolute requested-edit discrepancy from 1.979 to 4.44 \ensuremath{\times} 10\ensuremath{^{-16}} (Figure 6B). Thus, self-regression agreement was a constructed identity, while the separately specified oracle tested the intended quantity. The correction is explained by exact coordinate transport; no learned repair capability was tested.

\subsection{An unanchored defect required informative testing or explicit abstention}
Residuals for the two anchored axes remained 2.22 \ensuremath{\times} 10\ensuremath{^{-16}}, and no anchor residual is defined for the unanchored third axis. Empirical random-probe detection rates followed the specified law 1 \ensuremath{-} (2/3)\ensuremath{^{n}}, with maximum absolute deviation 0.0142 across the six budgets (Figure 7). One targeted complement probe detected the constructed sign defect. With only the original anchors, the scope gate withheld the third-direction interpretation. Failure to detect a discrepancy and absence of support for a named interpretation are therefore different outcomes; an unsupported direction was not assigned a zero effect.

\subsection{Tolerance choices exposed different failure modes}
The absolute and hybrid rules had low healthy rejection under the specified Gaussian noise, but weak sign-reversed responses could remain within their tolerance. The relative rule narrowed with \ensuremath{\delta} and could reject healthy noisy measurements when the reference response was small. At \ensuremath{\delta} = 10\ensuremath{^{-4}}, the healthy false-positive rates were 0.0000, 0.9590 and 0.0000, respectively. The corresponding defective detection rates were 0.0015, 0.9941 and 0.0015. The maximum empirical-versus-analytic rate discrepancy was 0.0049 (Figure 8). For the absolute 5\ensuremath{\sigma} rule, the Gaussian false-positive probability is 5.73 \ensuremath{\times} 10\ensuremath{^{-7}}; no rejection in a finite simulated cell does not make that probability zero. These outcomes describe sensitivity to an imposed sign defect under known noise; they do not establish a universal tolerance or clinical acceptance threshold.

\begin{table}[htbp]
\centering\small
\caption{Full-rank anchor calibration at measurement-noise SD 0.2}
\begin{tabular}{@{}p{0.288\linewidth} p{0.288\linewidth} p{0.288\linewidth}@{}}
\toprule
\textbf{Repeats} & \textbf{Inverse error} & \textbf{Procrustes error} \\
\midrule
1 & 0.324 (0.197--0.725) & 0.172 (0.060--0.354) \\
4 & 0.155 (0.090--0.279) & 0.086 (0.031--0.192) \\
16 & 0.084 (0.042--0.120) & 0.045 (0.013--0.075) \\
64 & 0.039 (0.025--0.064) & 0.019 (0.007--0.043) \\
\bottomrule
\end{tabular}
\par\medskip\raggedright\footnotesize Values are median RMS semantic mapping error (empirical 2.5th--97.5th percentiles) over 64 seeds. These are descriptive simulation intervals.
\end{table}
\section{Discussion}
The audit separates three questions that can otherwise share a single performance label: does a model predict the complete state, does a coordinate have a justified name, and does an edit correspond to the named perturbation? In the controlled linear setting, the first question can be answered identically under many bases while the latter two require explicit coordinate information. Exact output regrouping supplies a useful implementation baseline before attributing any benefit to a new factor architecture. The counterexample concerns a restricted information setting and does not contradict learned neural gains or establish a general impossibility theorem.

For vascular research, anchor design should specify observable perturbation signatures rather than assign ecology labels to factor indices. A controlled experimental program could vary loading or shear conditions, cellular conditions, and physicochemical--immune conditions, then estimate the joint response signatures with appropriate controls. Such interventions may affect more than one ecology: rank should be estimated from the resulting signatures, not inferred from the number of intervention labels. The FLOW framework offers a possible platform context for structured vascular experiments \cite{ref9}, but no chip experiment was performed here. Whether sufficiently informative, reproducible and independent signatures can be measured remains an empirical question.

A practical factor report could accompany a named interpretation with the anchor matrix, its estimated span, smallest retained singular value, measurement variability, held-out response checks and a model-version identifier. A proposed change outside the calibrated span should retain its unsupported component as unknown. Changes in encoders or projector coordinates should prompt transport verification or recalibration, even when predictive readout remains stable. For nonlinear representations, local perturbation Jacobians and local coverage might replace the global linear map, but that extension requires its own analysis and experiments.

Clinical anchoring requires a further evidential layer. A simulator state edit, an experimental perturbation response and a patient treatment contrast are different estimands. The observational IABP study motivates this distinction \cite{ref4}; its reported association is not an intervention oracle. Neither semantic alignment nor low prediction error resolves treatment timing, confounding or transport to clinical populations. Prospective or otherwise justified causal evidence would be needed before using factor edits for patient treatment planning.

The expanded controls make the distinction between execution and scientific acceptance operational. ScienceIDE separates valid tasks, defective baselines and numerical acceptance criteria \cite{ref10}. Our finite fault bank applies that distinction to coordinate reconstruction, prediction, delivered edits and supported scope. A prediction-preserving representation can still support an incorrect named edit, and a local calibration residual can omit errors introduced elsewhere in the pipeline. Testing the end-to-end physical quantity is therefore necessary for the particular interpretation being claimed. Consistent changes of latent units provide a useful valid control: a verifier that confuses its chosen coordinate convention with the physical quantity can itself reject a correct pipeline.

The probe experiment also exposes a limit of validation by repetition. Additional tests improve detection only when they can sample the affected direction. Repeated checks confined to the original anchor span cannot reveal a defect restricted to its complement. Random testing has a calculable coverage probability in this constructed setting; a targeted complement probe has a different information budget. Reporting these budgets prevents the targeted result from being presented as performance under the original anchor-only boundary. Where informative probes are unavailable, abstention preserves the limit of the evidence without implying that the unknown response is zero.

ScienceBuddy distinguishes diagnostic interaction feedback from the separately evaluated trajectory reward under fixed rubric and evaluator parameters, and retains versions and acceptance decisions during procedural revision \cite{ref11}. This motivates recording what was changed, what each check measured and why a candidate was accepted. Our fixed correction trace supplies such a record for one coordinate error, but it is not an experiment on harness adaptation, reinforcement learning or researcher collaboration. Its successful correction follows from a supplied mathematical rule. Establishing learned repair or persistent improvement would require separately defined learners, resource-matched comparisons and held-out tasks.

DFM separates externally validated progress in the present investigation from improvement in future discovery behavior, while emphasizing simulator dependence and provenance \cite{ref12}. The present results support algebraic checks and behavior under the declared simulation assumptions. They do not measure future discovery capability or transfer to clinical research. The apparent independence of the oracle is also specific: it avoids a deliberately shared code defect, while retaining the same stated mathematical world. Independent implementation, independent biological measurements and independent clinical validation are different sources of evidence.

The added experiments increase the number of tested dimensions, selected defects and verification conditions, but do not expand the biological validity of the simulator. The fault classes are specified examples, not a representative sample of all errors. The probe law assumes uniform axis selection and a single affected direction. The tolerance experiment assumes a known reference scale, independent Gaussian measurement noise and one sign-reversal defect. The hybrid rule avoids an excessively narrow tolerance at small signals in this setting, but cannot recover information when the response is too weak relative to noise. Reference agreement remains conditional on the reference and observables being scientifically appropriate; no verifier demonstrated here supplies a patient treatment oracle.

The main limitations are intentional. The core three arbitrary coordinates, additional dimensional controls, linear dynamics, identity action effects, Gaussian measurements and a known orthogonal prior make the audit transparent but restrict its scope. The persistence coefficients are not estimated from biology. We do not train an encoder, evaluate multimodal clinical sequences, compare learned regularizers, or reproduce any published neural benchmark. Our exact-span result does not incorporate intervention misspecification, nonlinear response, heterogeneous patients or hidden confounding. Temporal structure and additional identifying priors can alter the ambiguity result \cite{ref6,ref7}. The contribution is a reproducible translation of established coordinate and calibration mathematics into a vascular-motivated audit, with explicit abstention, conditioning controls and executable fault contracts; it is not a new foundation model or a validated medical device.

Within this boundary, complete prediction and semantic intervention can be tested separately. Known coordinate transport, informative anchors and stable calibration support named interpretation in the declared model; absent those conditions, factor geometry alone provides insufficient evidence. Future vascular digital-twin evaluations should test these links using independent measured responses before making mechanism or clinical decision claims.

\section{Mathematical appendix}
\subsection{Exact span recovery}
For B = Q\ensuremath{^{\mathsf T}}A and orthogonal Q, \ensuremath{B^{\dagger}=A^{\dagger}Q}. Therefore \ensuremath{AB^{\dagger}Q^{\mathsf T}=AA^{\dagger}=P}, proving Equation (3). Likewise \ensuremath{QBA^{\dagger}=P}. Every orthogonal synthesis map R satisfying RB = A agrees on span(B). If two such maps differ by an orthogonal transformation acting only on its complement, the anchor observations cannot distinguish them. In the core three-dimensional system, the complement has \ensuremath{O(3-\operatorname{rank}(A))} freedom; in rank two this includes a sign reflection when handedness is unspecified.

\subsection{A finite-noise bound for the inverse map}
For square invertible A, let \ensuremath{\alpha} denote its smallest singular value and let \ensuremath{\widetilde B=Q^{\mathsf T}A+E}. If \ensuremath{\lVert E\rVert_2<\alpha}, then \ensuremath{\widetilde{B}} is invertible. The inverse calibration \ensuremath{T=A\widetilde B^{-1}} satisfies \ensuremath{T-Q=-QE\widetilde B^{-1}}. Hence

\begin{equation}
\lVert TQ^{\mathsf T}-I\rVert_2\leq\frac{\lVert E\rVert_2}{\alpha-\lVert E\rVert_2}.
\tag{4}\label{eq:4}
\end{equation}
This deterministic bound explains the loss of stability when perturbation noise approaches the weakest signature scale. It applies to the unrestricted square inverse under the stated norm condition, not to every pseudoinverse or to the Procrustes estimator. A numerical rank declaration alone cannot guarantee a small denominator.

\section{Declarations}
\subsection{Data and code availability}
The arXiv ancillary archive contains seeds, specifications, executable analysis and audit scripts, seed-level results and original figures. The supplied generation scripts recreate all synthetic observations; serialized observation arrays are omitted from this compact public archive. The data are simulated and contain no identifiable or clinical information. No restricted patient dataset from the cited studies was obtained.

\subsection{Ethics and consent}
No human participants, animals or identifiable personal records were used. No ethics approval or participant consent was obtained for this purely mathematical and synthetic computational study.

\subsection{Funding}
This work was supported by the National Natural Science Foundation of China (T2288101 and 82170342), the Medical Engineering Joint Fund of Fudan University (yg2023-01), and the AI for Science Foundation of Fudan University (FudanX24AI003).

\subsection{Conflicts of Interest}
The authors declare no competing interests.

\clearpage
\begin{figure}[p]
\centering
\includegraphics[width=\linewidth,height=0.72\textheight,keepaspectratio]{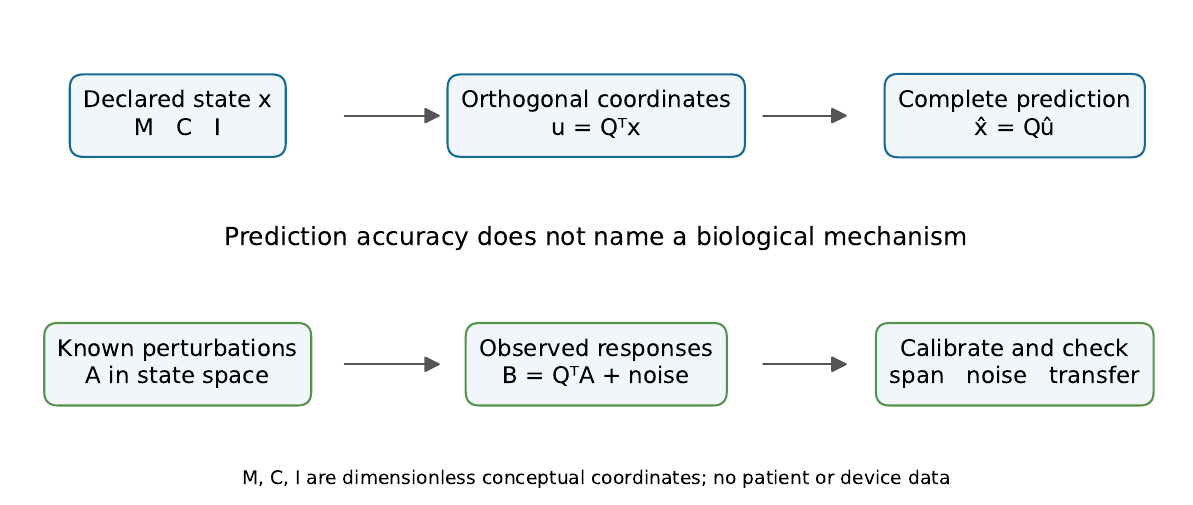}
\caption{\textbf{Audit workflow}. The upper path represents complete orthogonal analysis and synthesis. The lower path adds known perturbation signatures to calibrate interpretation. The three dimensionless state coordinates are conceptual analogues, not measured biological quantities. The audit evaluates coverage, noise sensitivity and coordinate transfer.}
\label{fig:Figure1_audit_workflow}
\end{figure}
\begin{figure}[p]
\centering
\includegraphics[width=\linewidth,height=0.72\textheight,keepaspectratio]{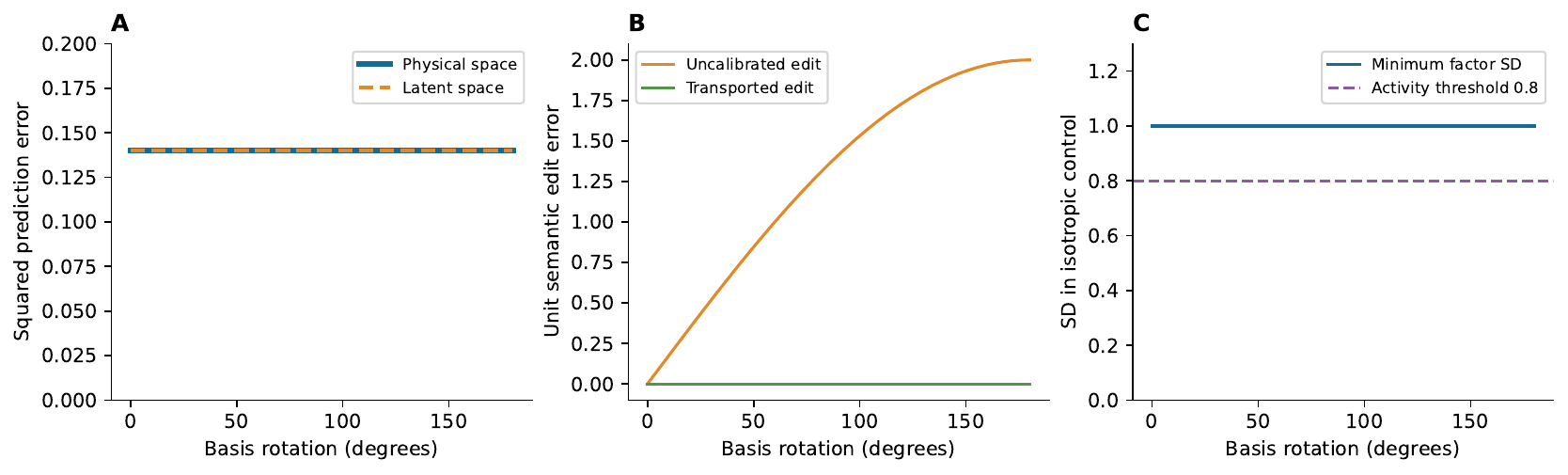}
\caption{\textbf{Prediction and semantic edits under basis rotation}. A, the squared norm of the fixed residual (0.2, \ensuremath{-}0.1, 0.3) remains 0.14 in either space. B, the error of an uncalibrated unit edit in the first coordinate increases with a rotation in the first two coordinates; transported edits recover the intended result. C, all factor SDs remain one for the separately constructed isotropic tight-frame control. This activity result is restricted to that control and does not hold for general covariance.}
\label{fig:Figure2_rotation_controls}
\end{figure}
\begin{figure}[p]
\centering
\includegraphics[width=\linewidth,height=0.72\textheight,keepaspectratio]{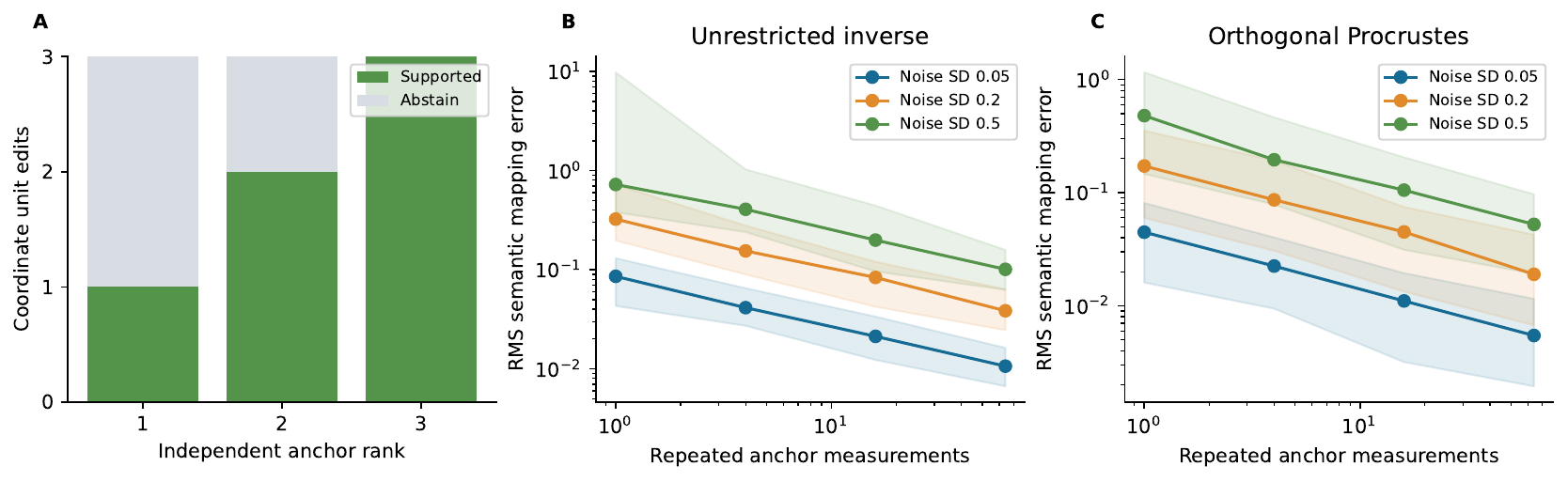}
\caption{\textbf{Anchor coverage and measurement noise}. A, k independent coordinate anchors support k of three specified unit edits; unsupported components require abstention. This counting example uses orthogonal coordinate anchors. B--C, full-rank calibration errors for the unrestricted inverse and orthogonal Procrustes map, respectively. Curves show medians across 64 seeds; shaded bands show empirical 2.5th--97.5th percentiles. Increasing repeat counts are nested within each seed. Bands are descriptive simulation variation, not clinical confidence intervals.}
\label{fig:Figure3_anchor_span_and_noise}
\end{figure}
\begin{figure}[p]
\centering
\includegraphics[width=\linewidth,height=0.72\textheight,keepaspectratio]{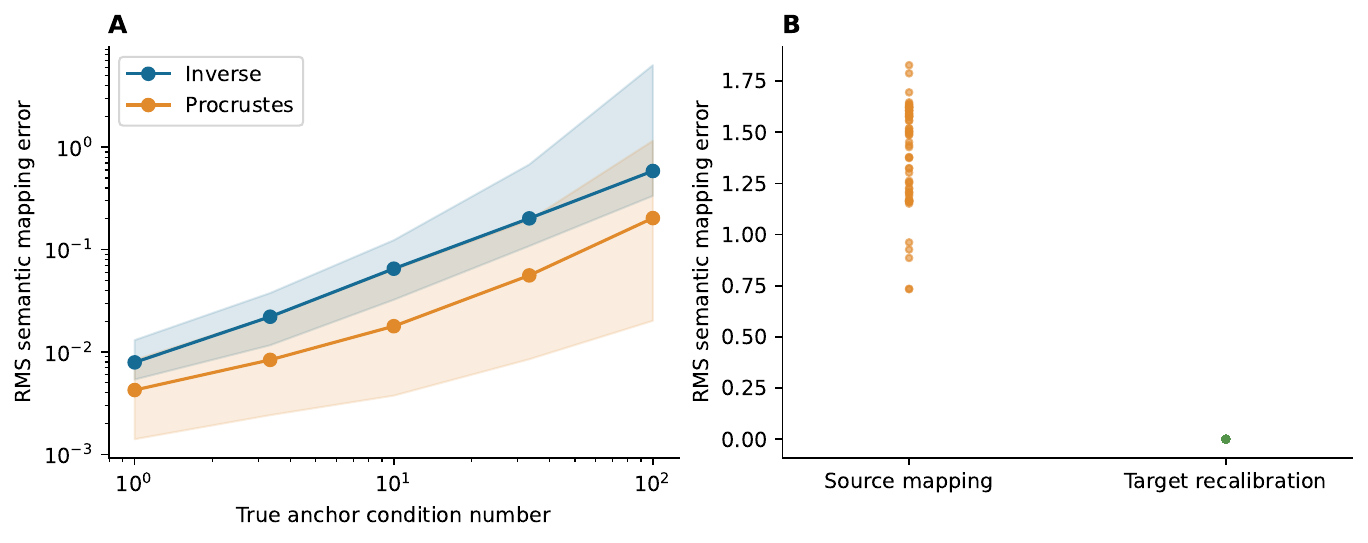}
\caption{\textbf{Conditioning and coordinate transfer}. A, full-rank anchors A = diag(1,s,s) become weaker as the true condition number increases. Measurement-noise SD is 0.02, averaged over 16 repeats. Median and empirical percentile bands summarize 64 seeds. B, each point denotes one independently sampled source--target basis pair. A source semantic map fails under target coordinate changes; noiseless target recalibration restores consistency. This test does not represent physiological domain shift.}
\label{fig:Figure4_conditioning_and_transfer}
\end{figure}
\begin{figure}[p]
\centering
\includegraphics[width=\linewidth,height=0.72\textheight,keepaspectratio]{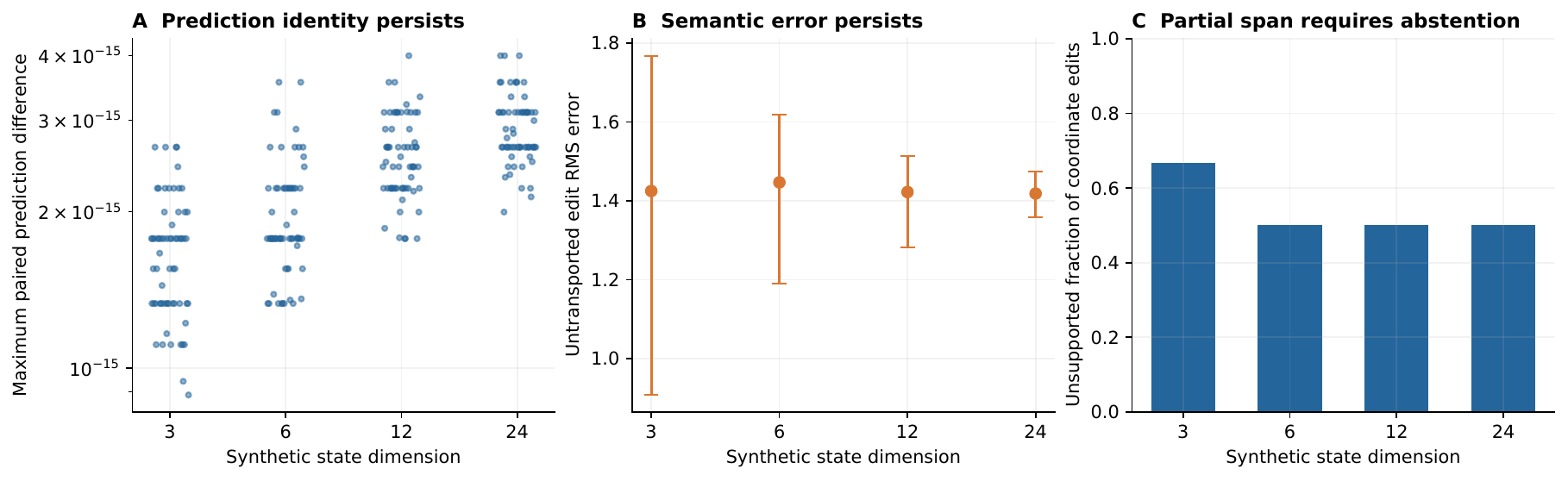}
\caption{\textbf{Complete coordinates across synthetic dimensions}. A, paired prediction discrepancies for 64 seeds in each of d = 3, 6, 12 and 24, using 256 training and 128 independent test observations in every dimension. Both predictors have d(2d + 1) parameters and d scalar output heads. B, untransported edit error; points are medians and bars empirical 2.5th--97.5th percentiles. C, unsupported fraction of d unit-coordinate edits with \ensuremath{\lfloor}d/2\ensuremath{\rfloor} coordinate anchors. The supported-span identity was checked separately in each replicate. These are implementation stress tests, not measured vascular mechanisms.}
\label{fig:Figure5_dimension_stress}
\end{figure}
\begin{figure}[p]
\centering
\includegraphics[width=\linewidth,height=0.72\textheight,keepaspectratio]{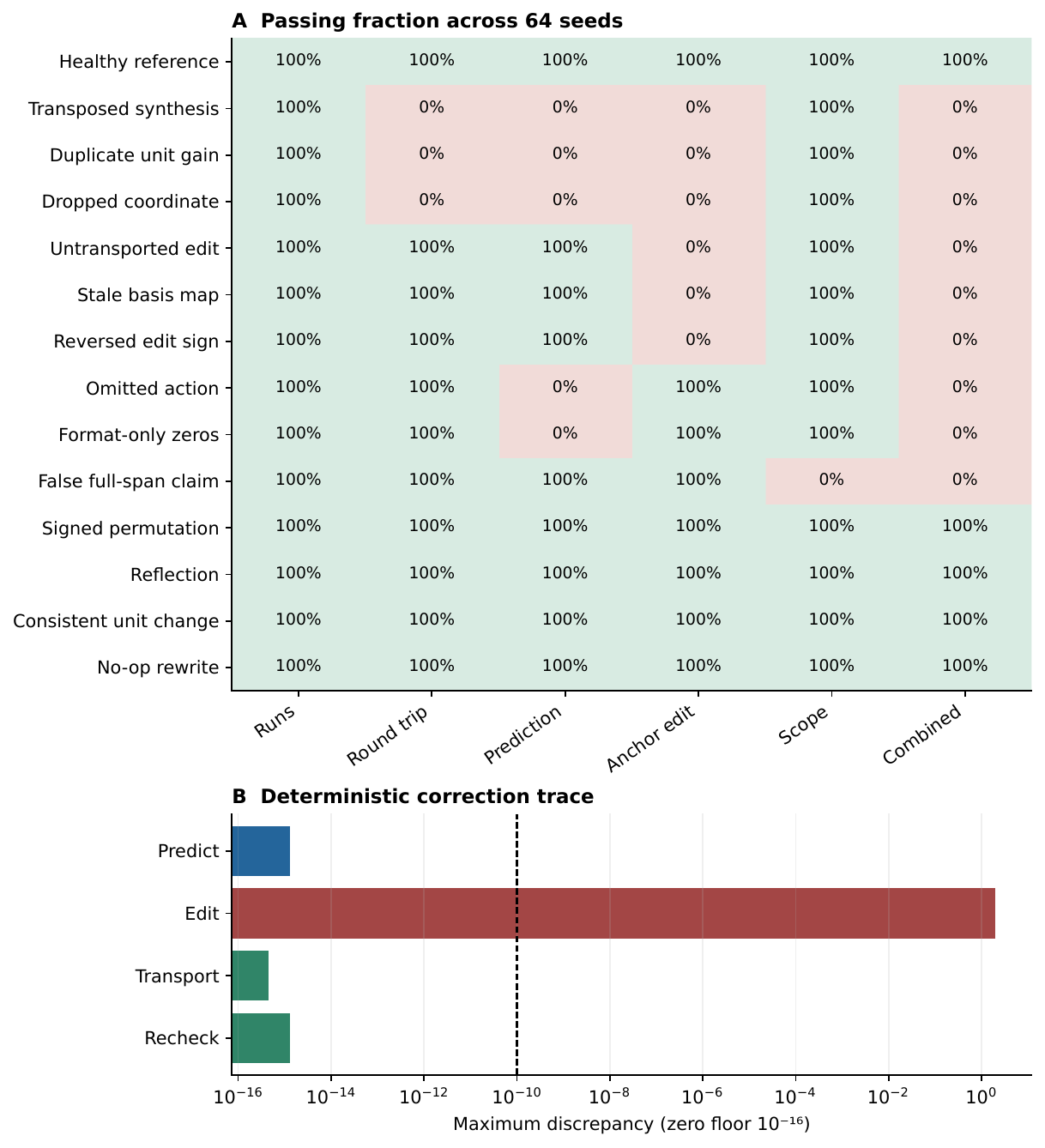}
\caption{\textbf{Executable contracts and deterministic correction}. A, passing fractions across 64 seeds for 14 runnable variants. Nine faulty variants give 576 faulty records; the healthy reference and four valid alternatives give 320 valid records. Five acceptance contracts assess execution, named-space round trip, noiseless prediction, actual delivered anchor edits and declared scope. The combined check requires all five. Consistent unit changes preserve physical outputs despite scaled latent coordinates. Another 64 execution-crash controls are recorded separately and omitted from the matrix. B, seed-zero untransported-edit control: prediction passes, the actual edit fails, prescribed coordinate transport restores the edit, and all checks pass on re-execution. The dashed line is the 10\ensuremath{^{-10}} noiseless tolerance; plotting uses a floor of 10\ensuremath{^{-16}}. This is a deterministic correction, not a learned agent repair. Fault counts describe a constructed bank, not natural error prevalence.}
\label{fig:Figure6_verifier_contracts}
\end{figure}
\begin{figure}[p]
\centering
\includegraphics[width=\linewidth,height=0.72\textheight,keepaspectratio]{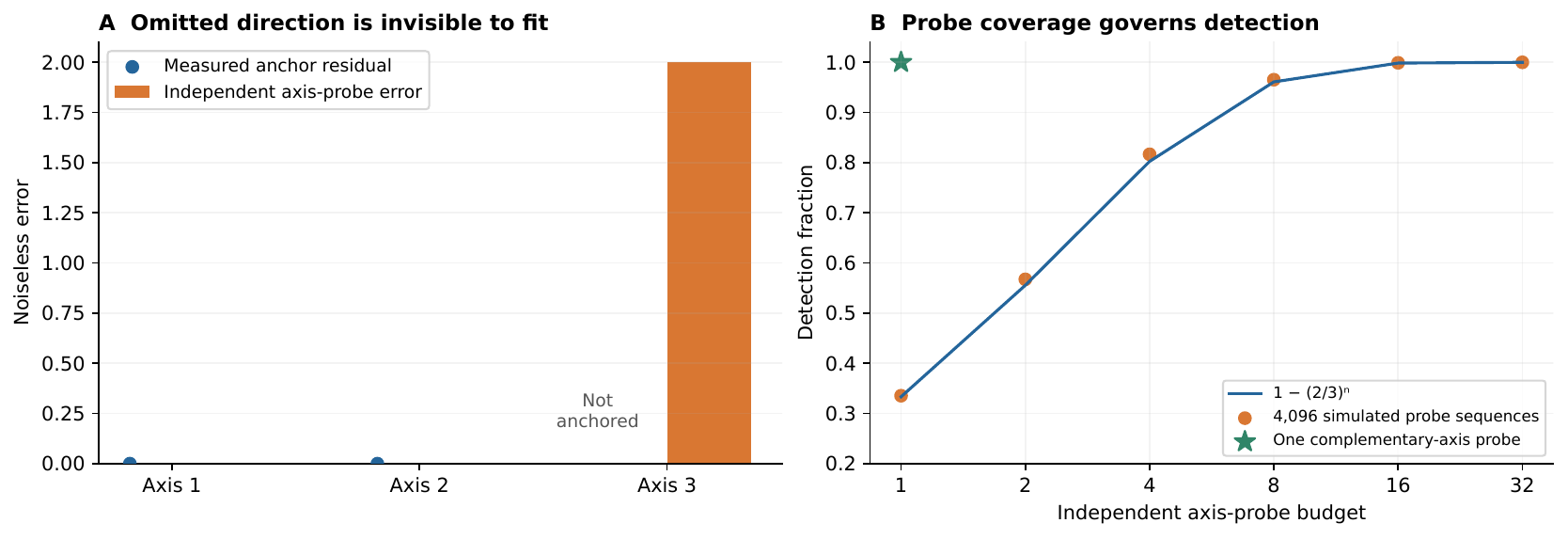}
\caption{\textbf{Independent probes of an unsupported direction}. A, a sign reversal in the third direction leaves the measured rank-two anchor residuals near zero; the third direction is not anchored and has no fit residual. Independent axis probes reveal an error of two there. B, 4,096 nested random probe sequences sample axes uniformly with replacement. Detection follows 1 \ensuremath{-} (2/3)\ensuremath{^{n}} for this single faulty axis. A single complementary-axis probe is a deterministic control with additional information. The original rank-two scope gate can withhold third-direction interpretation without detecting the defect. These probe budgets describe known noiseless simulator responses, not clinical detection rates.}
\label{fig:Figure7_independent_probes}
\end{figure}
\begin{figure}[p]
\centering
\includegraphics[width=\linewidth,height=0.72\textheight,keepaspectratio]{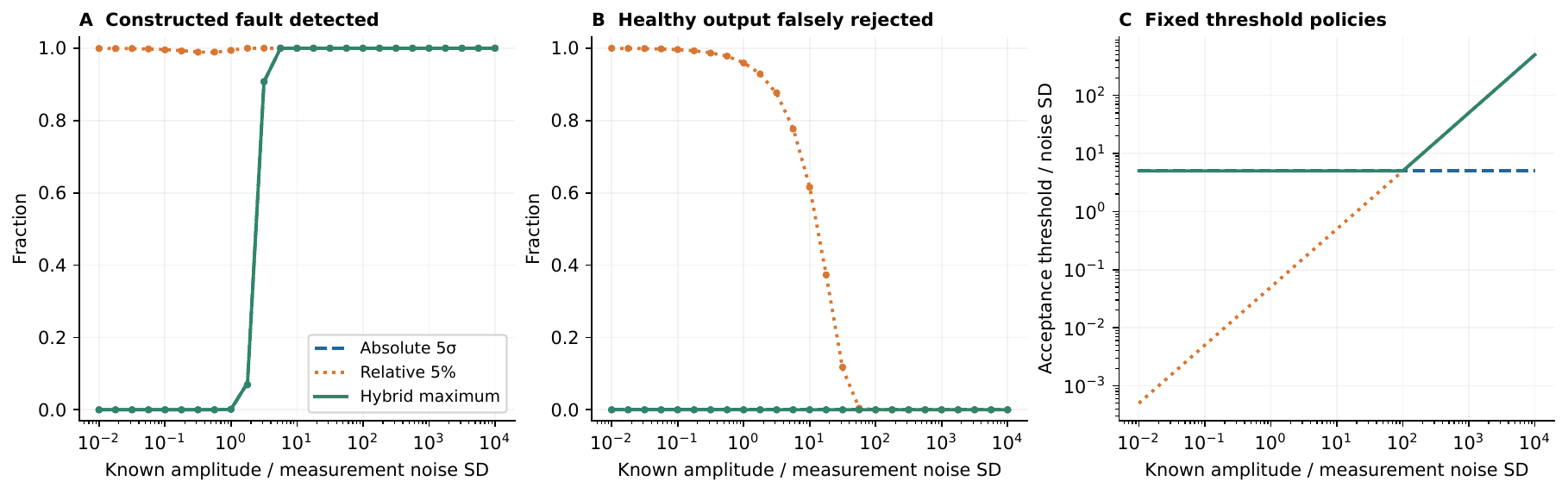}
\caption{\textbf{Acceptance tolerance under scalar measurement noise}. A, rejection of the constructed sign-reversal fault. B, erroneous rejection of healthy outputs. Healthy measurements are \ensuremath{\delta} + \ensuremath{\varepsilon} and faulty measurements \ensuremath{-}\ensuremath{\delta} + \ensuremath{\varepsilon}, compared with reference \ensuremath{\delta}, with scalar \ensuremath{\varepsilon} \ensuremath{\sim} N(0,10\ensuremath{^{-8}}). At each of 25 \ensuremath{\delta} values from 10\ensuremath{^{-6}} to 1, 16,384 measurements per condition evaluate \ensuremath{\tau} = 5\ensuremath{\sigma}, \ensuremath{\tau} = 0.05\ensuremath{\delta} and \ensuremath{\tau} = max(5\ensuremath{\sigma},0.05\ensuremath{\delta}). Lines show analytic Gaussian probabilities and points the Monte Carlo rates. C, the three fixed thresholds relative to \ensuremath{\sigma} = 10\ensuremath{^{-4}}. Coincident rules yield overlapping curves. The known response scale is not a clinical effect threshold; weak faults can be missed and narrow relative tolerances can reject healthy measurements.}
\label{fig:Figure8_tolerance_sensitivity}
\end{figure}
\clearpage
\end{document}